\RequirePackage{fix-cm}
\documentclass[twocolumn]{svjour3}          % twocolumn
\smartqed  % flush right qed marks, e.g. at end of proof
\usepackage{graphicx}
\journalname{IJCV}

\usepackage{amsmath}
\usepackage{times}

\begin{document}

\title{
Data-Dependent Higher-Order Clique Selection for Artery-Vein Segmentation by Energy Minimization
}

%\subtitle{Do you have a subtitle?\\ If so, write it here}

\titlerunning{Data-Dependent Higher-Order Clique Selection}        % if too long for running head

\author{
		Yoshiro Kitamura \and
		Yuanzhong Li     \and
		Wataru Ito       \and
		Hiroshi Ishikawa
}

%\authorrunning{Short form of author list} % if too long for running head

\institute{Y. Kitamura, Yuanzhong Li, Wataru Ito \at
              Imaging Technology Center, Fujifilm Corporation, Tokyo, Japan \\
              \email{yoshiro.kitamura@fujifilm.com}           %  \\
           \and
           Y. Kitamura, H. Ishikawa \at
              Department of Computer Science and Engineering, Waseda University,
              Tokyo, Japan, and JST CREST
}

\date{Received: date / Accepted: date}
% The correct dates will be entered by the editor

\maketitle

\begin{abstract}
We propose a novel segmentation method based on energy minimization of higher-order potentials. We introduce higher-order terms into the energy to incorporate prior knowledge on the shape of the segments. The terms encourage certain sets of pixels to be entirely in one segment or the other. The sets can for instance be smooth curves in order to help delineate pulmonary vessels, which are known to run in almost straight lines. The higher-order terms can be converted to submodular first-order terms by adding auxiliary variables, which can then be globally minimized using graph cuts. We also determine the weight of these terms, or the degree of the aforementioned encouragement, in a principled way by learning from training data with the ground truth. We demonstrate the effectiveness of the method in a real-world application in fully-automatic pulmonary artery-vein segmentation in CT images.

% \PACS{PACS code1 \and PACS code2 \and more}
% \subclass{MSC code1 \and MSC code2 \and more}
\end{abstract}

\section{Introduction}
\label{intro}
Energy minimization is regularly used for image labeling problems such as segmentation and stereo. Higher-order energies are perhaps not as common, but are nevertheless being used more and more (Wang et al. 2013). Whereas the common first-order (pairwise) potential can directly model only the relationship between pairs of pixels, the higher-order potential can model more complex and useful relationships between more than two variables. In this paper, we present a novel segmentation method based on higher-order energy minimization where we choose the higher-order cliques according to the data. We demonstrate the efficacy of this method in a real-world application in medical imaging. Specifically, we present a fully-automatic pulmonary artery-vein segmentation in 3D CT images, which is hard since the artery and the vein have similar CT values, making them indistinguishable locally, and are often entwined with each other in the lung, often with a large contacting area.

\noindent
{\bf Data-Dependent Higher-order Clique Potentials}\hspace{12pt} A significant issue in using higher-order potential is the choice of the cliques, i.e., the variables on which each potential term is defined. In the case of the pairwise potentials, a clique is usually a pair of neighboring variables and the potential regularizes (i.e., smoothes) the labeling. Similarly, in the higher-order energies, the cliques are often taken as predetermined regular shapes, typically fixed blocks such as squares (Kohli et al. 2009a, Russell et al. 2007, Ishikawa 2011). Kohli et al. (2009b) used as cliques superpixel-like image patches of arbitrary shapes, generated by unsupervised segmentation algorithms such as mean-shift. Recently, Kadoury et al. (2013) used a similar approach. Again, these higher-order potentials regularize the labeling, encouraging the homogeneity and regional consistency.

In this paper, we consider a different use of higher-order potentials: to incorporate the prior knowledge on the shape of the segments into the energy. Prior knowledge regarding the shape of the segments is crucial in difficult segmentation problems, as it is often the only feature that delineates one segment from another. To exploit it, we encourage the segmentation that is more in accordance with the prior knowledge. In the case of the labeling representation of segmentation, the shape is expressed by the set of pixels that constitute the shape. Therefore, to encourage certain shapes, we encourage specific set of pixels to have the same label. However, since the number of combinations of pixels in a higher-order clique increases exponentially with the order, using all possible combinations of pixels and defining potentials according to the likelihood of each is prohibitively inefficient. Thus, some restriction is needed to effectively choose useful cliques. Here, we propose to adaptively choose the cliques according to the given data and the prior knowledge about what kind of shape is desired for the given segmentation problem. We utilize the (robust) $P^n$ Potts potential (Kohli et al. 2009a, 2009b) to encourage the pixels in each cliques chosen thus to have the same label. The $P^n$ Potts potential is a higher-order potential that can be converted to an equivalent submodular first-order (pairwise) potential, thus allowing efficient global optimization.

Learning the potential function is another important aspect in utilizing higher-order potentials. Even in the first-order case, learning the parameters of the potentials significantly improves the results (Scharstein and Pal 2007, Sun et al. 2008, Li 2009); it is even more important in the higher-order case, since it becomes increasingly more difficult to design the potential by trial and error, as often is done in practice. Thus, we also learn the parameters of the potential from the training data. Moreover, we also learn the parameter for the adaptive choice of the cliques according to the given data.

\noindent
{\bf Related Work}\hspace{12pt} Many conventional segmentation methods utilize length-based regularizer represented by first-order potentials. Its main drawback is a shrinking bias that often leads to miss-segmentation of thin and elongated structures like vessels. One approach to avoid the bias is using curvature-based measures (El-Zehiry and Grady 2010, Schoenemann et al. 2012, Standmark et al. 2011). Curvature regularizer generally require non-submodular and/or higher-order potentials, complicating the optimization process. Discretization of curvature angles enables efficient computation but causes discretization artifact. In a similar manner, Olsson et al. (2013) transform curvature regularization into a multi-label optimization with patch based variables. These methods have a similar tradeoff between the computation costs and accuracies. Recent work of Nieuwenhuis et al. (2014) models efficient squared curvature, computational costs of which increases linearly as the angular resolution increases.

Shekhovtsov et al. (2012) learn local patterns and costs of curvature which work like filters whose response is locally minimized. The learned potentials are thought to be optimal for the assumed application. The method consists of exclusively selecting a best pattern for each local window. However, its computational cost is not small enough for real time application. Our work is similar to their work in that we also learn patterns and weights from training samples, but is different in adaptive sampling of variables in cliques. This enables us to achieve further computational efficiency.

Another approach to segment thin and elongated structures is using connectivity constraints. Nowozin and Christoph (2009) derived a linear programming relaxation that enforces the output labeling to be connected in Markov random field models. St\"uhmer et al. (2013) (later extended by Oswald et al. 2014) present the tree-shaped connectivity prior that is constructed by a geodesic shortest path tree. They include the prior in an efficient convex optimization framework. The expression of connectivity by shortest path tree in St\"uhmer is similar to our method, but the embodiment in an energy function is different as ours are based on higher-order submodular potentials. We consider that our method has an advantage in reducing the size of optimization problem via adaptive clique selection procedure.

\begin{figure}
\begin{center}
\includegraphics[width=80mm]{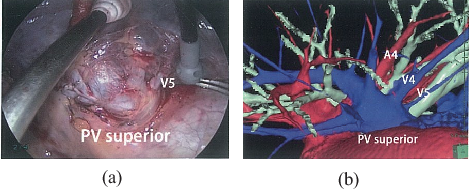}
\end{center}
\caption{The impact of pre-surgery simulator. (a) A thorascopy image, (b) 3D rendering from the viewpoint matching the thorascopy image. We can grasp the 3D positions of the branches such as pars medialis (V5), ramus lateralis (A4), and pars lateralis (V4) from the 3D rendering image (b). In contrast, any structure behind V5 cannot be observed in the thorascopy image (a). Referring to the 3D rendering images before and during a surgery can improve the accuracy and safety of video-assisted thorascopic surgery.}
\label{figSurgery}
\end{figure}

\noindent
{\bf Pulmonary Artery-Vein Segmentation}\hspace{12pt} Lung cancer is the most common cause of cancer-related deaths in the world (Ferlay et al. 2010). Thanks to the multi-detector CT, which has become common in clinical practice, lung cancers can now be detected in early stages, while minimally invasive surgery such as video-assisted thorascopic surgery (VATS) lobectomy or segmentectomy can still be effective. The number of such surgeries is rapidly increasing in recent years. However, segmentectomy is a highly complex operation because of the high variability of vascular and bronchial structures, on top of the difficulties caused by the limited view and operability of a thorascopy. Therefore, pre-surgery simulation and navigation systems have great clinical importance (Ikeda et al. 2013), providing the precise knowledge of the anatomy of the patient's pulmonary vessels and bronchi as shown in Fig. \ref{figSurgery}. These systems in turn require each organ to be segmented in the CT images. Since manual segmentation is too time-consuming, precise and fully-automatic segmentation is the key to successful deployment of such systems. 

In this paper, we present a fully-automatic algorithm for segmentation of pulmonary arteries (PA) and veins (PV) from CT Angiography data, based on higher-order energy minimization. Note that typical graph-cuts methods utilizing first-order potentials tend to fail in PA/PV segmentation, since its boundary is locally indistinct, often with large contacting area. To overcome this challenge, the proposed method utilizes $P^n$ Potts potentials with cliques selected depending on the data. In the method, we encourage voxels on a smooth curve to all belong to one segment (or, put another way, discourage the curve from crossing segment boundary), to improve the accuracy of segmentation of pulmonary vessels, which is known to run in an almost straight line. To the best of our knowledge, this is the first method that realizes fully-automatic PA/PV segmentation achieving clinically acceptable accuracies.

\vspace{8pt}
A preliminary version of this paper appeared as (Kitamura et al. 2013). The present paper has more detailed presentation of the segmentation method as well as extended by addition of an improvement of the method by incorporating spatial arrangement features (\textsection\ref{subsec_spatial}), a quantitative comparison with an existing method (\textsection\ref{subsec_quantitative}), and subjective tests in clinical settings (\textsection\ref{subsec_subjective}). We continue as follows. 
After preliminaries on higher-order energy minimization as well as on pulmonary vessel segmentation in Section \ref{sec_prelim}, we describe the data-dependent higher-order clique potential in Section \ref{sec_ASPP}. In Section \ref{sec_segdetail}, we discuss the details of the application in fully-automatic pulmonary artery-vein segmentation. We will show quantitative comparison with an existing method as well as subjective tests done by several clinicians in Section \ref{sec_experiments}. Finally, we discuss and summarize our method in Sections \ref{sec_discussion} and \ref{sec_conclusion}.

\section{Preliminaries}\label{sec_prelim}
%\label{sec:1}
%Text with citations \cite{RefB} and \cite{RefJ}.

\subsection{Higher-order energies, Pseudo-Boolean Functions, and Order Reduction}
The segmentation problem can be formulated as a pixel-labeling problem. Many conventional methods minimize the first-order energy function with unary and pairwise terms:
\begin{equation}
E(X) = \sum_{a \in V} \theta_a(x_a) + \sum_{a \in V, b\in N_a} \theta_{ab}(x_a, x_b),
\label{eqEnergyUandP}
\end{equation}
where $X$ is the vector of binary variables $x \in \{0, 1\}$ indexed by the set $V$ of pixels, and $N_a$ is the set of the neighbors of pixel $a$. The functions $\theta_a$ and $\theta_{ab}$ give the potential for the binary label $x_a$ and the label pair $(x_a, x_b)$, respectively. In this paper, we deal with higher-order energies, in which terms that depend on more than two variables appear.

%\label{sec:2}
%as required. Don't forget to give each section
%and subsection a unique label (see Sect.~\ref{sec:1}).
%\paragraph{Paragraph headings} Use paragraph headings as needed.
%\begin{equation}
%a^2+b^2=c^2
%\end{equation}

For a clique, or a subset $c$ of $V$, we denote the subset $\{x_a | a \in c\}$ of the variables by $X_c$.
 The energy \eqref{eqEnergyUandP} is a special case of:
\begin{equation}
E(X) = \sum_{c \in C} \theta_c(X_c),
\label{eqEnergyC}
\end{equation}
where $C$ is a set of cliques in the graph, and $\theta_c(X_c)$ are functions parameterized by $C$ that depend on $X$ only through the subset $X_c$ of the variables. The maximum of $|c| - 1$ among the cliques $c$'s is called the {\em order} of the energy. Thus, first-order energies have the cliques with at most two vertices in the sum, in which case the energy can be written in the form of \eqref{eqEnergyUandP}. In this paper, we consider more general form of energy, where the cliques can contain more than two vertices.

Note that a clique is usually defined as a complete subgraph. In this paper, we call any subset of $V$ a clique, as is common in the literature on higher-order energy minimization. We can always find a topology that makes given subsets of vertices to form a complete subgraph. Of course, in that case there would be many more cliques than the intended subsets, but we need not include them in $C$. Alternatively, we can use a constant as the function $\theta_c$ for them. Instead of having a fixed topology and therefore a fixed set of cliques, we adaptively choose which subset $c$ of vertices we use as the cliques in \eqref{eqEnergyC}. This departure from the notion of a predetermined set of cliques is a very important aspect of this work.

Throughout this paper, we take $V$ to be the set of pixels (or voxels) and the set of values to be $\{0,1\}$. In such a case, the energy \eqref{eqEnergyC} is a real-valued function of variables that take values in $\{0,1\}$. Such a function is called a pseudo-Boolean function. Pseudo-Boolean functions can always be written as a polynomial, e.g.,
\begin{align*}
\theta_{ab}(x_a, &x_b)=\\
&\theta_{ab}(0,0)(1 - x_a)(1 - x_b)+\theta_{ab}(0,1)(1 - x_a)x_b\\
&\hspace{40pt}+\theta_{ab}(1,0)x_a(1 - x_b)+\theta_{ab}(1,1)x_ax_b.
\end{align*}
In this quadratic polynomial form, if all of the weights of the quadratic terms in the expanded energy function are non-positive, it is {\em submodular} and can be globally minimized by the graph-cut algorithm in polynomial time (Hammer 1965). Otherwise, non-submodular functions can be partially optimized by the roof duality or QPBO algorithm (Hammer et al. 1984).

Recent advancements enable us to utilize higher-order energy functions, which contain, in the polynomial form, terms of degree higher than two. According to (Ishikawa 2011), a higher-order pseudo-Boolean function can be transformed into an equivalent first-order (quadratic) function by adding auxiliary variables. Then the original problem can be solved by minimizing the transformed function by conventional algorithms such as graph cuts or QPBO (Boykov and Kolmogorov 2004; Rother et al. 2007).

Submodular energy functions have great advantages since they can always be minimized globally in polynomial time using graph cuts. Thus, higher-order functions that are converted to submodular quadratic functions are of special interest. Two known cases are (Kohli et al. 2009a):
\begin{equation}
1-x_1x_2...x_n = 1+\min_{z \in \{0,1\}} -z(x_1+x_2+...x_n-n+1),
\label{eqOrderReduction1}
\end{equation}
\begin{align}
1-&(1-x_1)(1-x_2)...(1-x_n) = \nonumber\\
&\hspace{36pt}1+\min_{z \in \{0,1\}} (1-z)(x_1+x_2+...x_n-1),
\label{eqOrderReduction2}
\end{align}
where $z \in \{0, 1\}$ is the auxiliary variable and $n$ is the degree of the function.
The potential \eqref{eqOrderReduction1} takes the value $0$ if and only if its variables are all $1$ and otherwise takes value $1$.
Similarly, \eqref{eqOrderReduction2} takes the value $0$ only when its variables are all $0$ and otherwise takes value $1$.
These potentials are called the $P^n$ Potts model (Kohli et al. 2009a). Note that the right-hand side of the transformations are submodular quadratic pseudo-Boolean functions (i.e., the coefficients of all quadratic terms are non-positive) inside a minimization. 
From these, we can see that adding the higher-degree terms on the left-hand side to a minimization problem is equivalent to adding the quadratic function on the right-hand side and adding the auxiliary variable $z$ to the set of variables over which to minimize.

\begin{figure}
\begin{center}
\includegraphics[width=30mm]{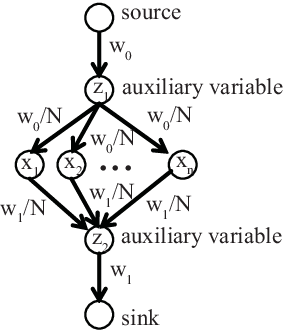}
\end{center}
\caption{The graph construction of the function: $w_0 \min \left( 1, \sum_{i=1,...,n}\frac{x_i}{N} \right)+w_1 \min \left( 1, \sum_{i=1,...,n}\frac{1-x_i}{N} \right)$ }
\label{figGraphConstruction}
\end{figure}

Although eqs. \eqref{eqOrderReduction1} and \eqref{eqOrderReduction2} take a smaller value only when all the variables have the same value (0 or 1), we can use the {\em robust} $P^n$ Potts model (Kohli et al. 2009b) which takes gradually larger value as the number of variables violating the condition increases. With a positive integer $N$, such potentials are formulated as:
\begin{equation}
\min \left(1, \sum_{i=1}^n \frac{1-x_i}{N} \right) = 1+\min_{z \in \{0,1\}}z\left(-1+\sum_{i=1}^n \frac{1-x_i}{N}\right),
\label{eqOrderReduction3}
\end{equation}
and
\begin{equation}
\min \left(1, \sum_{i=1}^n \frac{x_i}{N} \right) = \min_{z \in \{0,1\}}\left(z+(1-z)\sum_{i=1}^n \frac{x_i}{N}\right).
\label{eqOrderReduction4}
\end{equation}
They take the value 0 when the condition (all $1$ for \eqref{eqOrderReduction3} or all $0$ for \eqref{eqOrderReduction4}).
When some of the variables violate the condition, they take an increasing value until saturating at value 1 when $N$ or more variables violate the condition.
In terms of graph-cut construction, they correspond to the graphs such as the one illustrated in Fig. \ref{figGraphConstruction}.

\subsection{PA/PV Segmentation}
\begin{figure}
\begin{center}
\includegraphics[width=60mm]{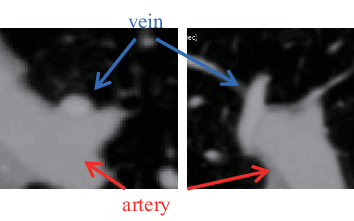}
\end{center}
\caption{Axial and sagittal images of the point of contact between the artery and the vein.}
\label{figContact}
\end{figure}

Pulmonary arteries (PA) and veins (PV) originate from the right ventricle and the left atrium, respectively, and branch several times passing through the hilum. In entirety, they have tree structures and are arranged radially in the lungs. Despite the progress of segmentation techniques, segmenting arteries from veins in CT images is still known to be difficult because of the following reasons: 
\begin{description}
\item[1)] PA and PV are indistinguishable from local images, because they have similar intensities and tubular shapes. 
\item[2)] They are often found entwined everywhere in the lungs, where the boundaries of the contact points are often unclear (Fig. \ref{figContact}).
\item[3)] The vascular structures are highly variable, depending on the patient.
\end{description}

Although considerable research has been done for extracting vessels, only a few methods have been published for separating arteries from veins. Early methods given by Lei et al. (2001) and van Bemmel et al. (2003) were applied to MR data, utilizing fuzzy connectedness or the level-set framework. More focused on pulmonary vessels, a preliminary study (Yamaguchi et al. 2002) utilizing an algorithm based on region growing has been reported. One promising method introduced by Saha et al. (2010) combines fuzzy distance transform and morphologic features. Later they incorporated the algorithm into a graphical user interface system, and showed high segmentation accuracy and reproducibility for non-angiograph images (Gao et al. 2012). Recently, Park et al. (2013) introduced an approach based on the minimum spanning tree algorithm that can handle the disconnected peripheral branches with clinically acceptable accuracies. However, these methods, which are designed for non-angiograph images, are interactive. That is, they require the user to provide numerous seed points on multiple branches. As for fully-automatic PA/PV segmentation, there has been one approach utilizing the specific anatomical knowledge: a pulmonary artery is often in close proximity to an airway, going in parallel (Mekada et al. 2006; Buelow et al. 2005). Mekada et al. demonstrated the effectiveness of the approach in a few cases where manually-labeled complete airway segmentations are available.

In the case of angiography images, the difficulty of the problem is eased as the origins of pulmonary vascular trees are visible on the image, though separating artery and vein remains difficult as they are often entwined with unclear boundaries. For angiography images, a fast-marching algorithm that propagates a front in the direction of minimal cost was used by Sebbe et al. (2003), while Ebrahimdoost et al. (2011) utilized a 3D level-set algorithm. However, these methods are only for segmenting pulmonary arteries, and in thorough evaluations for both of the artery and vein, they have not done well.

\section{Data-Dependent Clique Potential}\label{sec_ASPP}
We introduce a novel use of the higher-order energy minimization, which we call the {\em Data-Dependent Clique Potentials} (DDCP), in segmentation. The idea is to add (robust) $P^n$ Potts potential (the higher-order terms of the form \eqref{eqOrderReduction1},\eqref{eqOrderReduction2},\eqref{eqOrderReduction3}, and \eqref{eqOrderReduction4}) to encourage all of the $n$ variables in the term to have the same value (0 or 1). By choosing the variables (i.e., pixels) to include in such higher terms, we can encourage specific configurations of many variables while keeping the energy convertible to a submodular quadratic function. In other words, we can determine how much the pixel set is prone to being in the same segment. 

As a polynomial, each DDCP looks like \eqref{eqOrderReduction1} and \eqref{eqOrderReduction2}, i.e., it is just a $P^n$-Potts potential. The significance of an DDCP comes from the set of pixels it contains as variables. The idea is to utilize a prior knowledge regarding the shape in assigning pixels to potentials. By encouraging a set of pixels forming a certain shape to all belong to the same segment, we influence the shape of the resulting segments. 

In this paper, we consider a three-class labeling problem where the classes are {\it Background, Artery}, and {\it Vein}. We solve this in sequential two binary labeling steps. The first step separates the vessel regions from the background. The second step separates the vessel regions into artery and vein.

\begin{figure}
\begin{center}
\includegraphics[width=80mm]{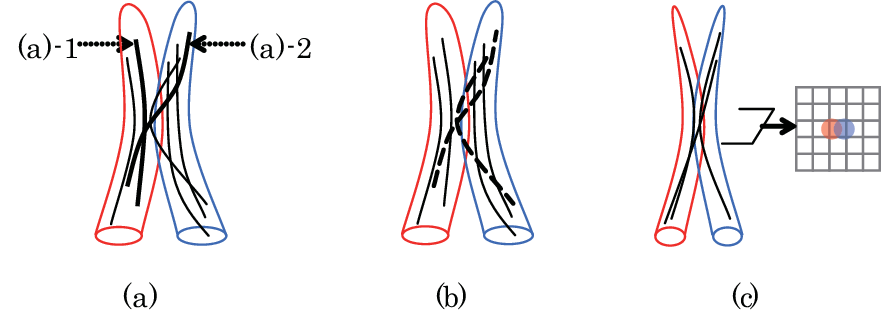}
\end{center}
\caption{Artery and vein are in contact in the middle. (a) The black curves illustrate the potentials. (b) After a successful separation. (c) A case where the image resolution is less than the vessel diameters; a sectional image is shown on the right.}
\label{figArteryVeinModel}
\end{figure}

For instance, we can encourage pixels forming a curve with low curvature to be in the same segment. Consider a segmentation problem of separating the artery (value $0$) and vein (value $1$) branches illustrated in Fig. \ref{figArteryVeinModel}(a), which is difficult using the first-order energy \eqref{eqEnergyUandP} as the two vessels are in contact. 
To introduce the DDCP in this case, we find relatively straight curves inside the vessels as shown in black in the figure, and form a potential from the voxels on each such curve, by adding a term of the form \eqref{eqOrderReduction1} or \eqref{eqOrderReduction2} or both to the energy. 
Adding \eqref{eqOrderReduction1} encourages all the variables (voxels) on the curve to take value $1$, i.e., encourage the curve to lie entirely in the artery segment.
Similarly, adding \eqref{eqOrderReduction2} encourages the curve to be in the vein segment.
If we add both, it encourage the curve to lie entirely in either artery or vein segment and discourage it from crossing the boundary.

The intention is that we would like each curve to fall entirely in one segment; but in general that cannot happen to all the curves, as in the case of this example, where some voxels belong to multiple curves that clearly belong to different segments. Then we modulate the degree of encouragement according to the curvature by making the positive coefficient multiplied to the potential larger for straighter curves (details can be found in \textsection\ref{subsec_implementation}). That way, we encourage straighter curves to be ``chosen,'' i.e., to be completely included in one of the segments. In the case of Fig. \ref{figArteryVeinModel}(a), the set (a)-1 of voxels on a straight line would be encouraged more than the (a)-2 that forms a curve. 

The key property of our method is that, even when these curves are overlapping, the segmentation can ``choose'' between them in this way. That is, as the result of segmentation, some of the curves are ``chosen'' and lay entirely in one segment, while others are ``unchosen'' and cross the segment boundary. Fig. \ref{figArteryVeinModel}(b) shows the successful result of such a choice, where the curves drawn with the solid lines are chosen and contribute to the decrease of the energy. It is important to note that the unchosen curves, drawn with the dashed lines, do not affect the resultant energy. As the higher order terms of the form \eqref{eqOrderReduction1} and \eqref{eqOrderReduction2} have an effect only if all the variables are labeled the same, the energy decreases only when the entirety of the curve is segmented as one. They are possible choices before the energy minimization, but once the segmentation is done and they are unchosen, {\em they do not cause any side effects}: they do not encourage any other unforeseen solution, as in the case of lower order potentials.
If, for instance, each voxel on the curve were encouraged individually, any number of other curves containing some of the voxels would be encouraged to varying degree, leading to an unpredictable behavior.
In an actual segmentation, the entire image has many points of contact that forces a choice as above, and the number of combinations of these choices is very large. However, graph cuts can find the best separation between PA and PV trees in terms of the energy.

To be sure, there is a theoretical limitation when the image resolution is close to the vessel diameters, as shown in Fig. \ref{figArteryVeinModel}(c). Here, the part the two vessels touch has the width of only about one voxel. In this case, the ``choosing'' of one curve would make inactive {\em all} of the higher-order terms that belong to the other vessel, rendering the effect of the DDCPs null. One way to ease this limitation is using the robust $P^n$ Potts model in eqs. \eqref{eqOrderReduction3} and \eqref{eqOrderReduction4}, instead of the original $P^n$ Potts model in \eqref{eqOrderReduction1} and \eqref{eqOrderReduction2}. They tolerate at most $N$ voxels having different labels, while still having the effect of decreasing the energy. However, even this model cannot handle it if the segments overlap too much, sharing many voxels. This also happens sometimes when two vessels run side by side in close proximity. 

Crucial to the effective use of the method is choosing which set of voxels to add in a principled way, adapting to the given data. We illustrate this process in more details in the case of the segmentation problem of pulmonary artery and vein in the following sections.

\section{Fully-automatic pulmonary artery-vein segmentation}\label{sec_segdetail}
To demonstrate the proposed method, here we present a fully-automatic method for segmentation of chest CT Angiography data. The CTA data is acquired after injecting contrast agents into patients. Therefore, the contrast of arteries and veins is enhanced from their roots to the peripheral branches, making it possible to follow the vessels from the roots to the peripheral. 

\subsection{Pulmonary vessel segmentation}
\begin{figure}
\begin{center}
\includegraphics[width=80mm]{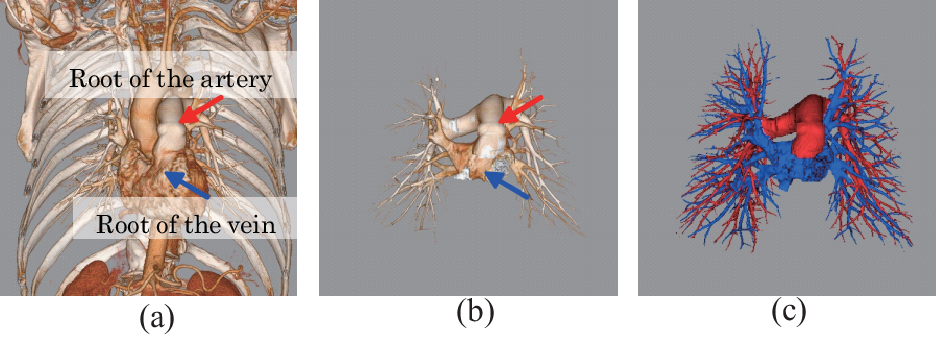}
\end{center}
\caption{Schematic images of the pulmonary artery-vein segmentation. The method consists of the three steps: (a) root-position detection, (b) vessel-region extraction, and (c) artery-vein segmentation. (reprinted from Kitamura et al. 2013)}
\label{figAVSegmentation}
\end{figure}

The segmentation method we describe here consists of the following three steps: root position detection, vessel region extraction, and artery-vein separation.
Schematic images of the steps are shown in Fig. \ref{figAVSegmentation}.

\begin{figure}
\begin{center}
\includegraphics[width=60mm]{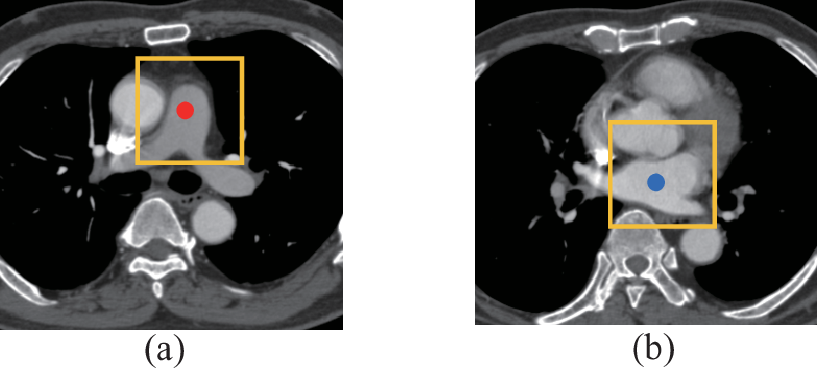}
\end{center}
\caption{Example images of the learned detector: (a) the pulmonary artery trunk and (b) the left atrium of the heart. The yellow rectangles represent the window size of the detectors.}
\label{figAppearanceOfRoots}
\end{figure}

\noindent
{\bf Root position detection}\hspace{12pt} The root of the pulmonary arteries and veins are the pulmonary artery trunk and the left atrium of the heart, respectively. These can be detected by specialized landmark detectors (Wang et al. 2009). In our implementation, two types of appearances on axial images were learned from training data by using machine learning (Friedman et al. 2000). Fig. \ref{figAppearanceOfRoots} shows the examples of the learned appearances. Each root position in the input data is detected by scanning it by the learned detector.

\noindent
{\bf Vessel region extraction}\hspace{12pt} The vessel regions are segmented by the conventional graph-cut method which utilizes unary and pairwise potentials. Pulmonary vessels have different characteristics in the mediastinum and the lung. The thick vessels in the mediastinum are extracted as continuously extending regions from the detected root positions. To do this, foreground seeds (unary terms) are set around voxels where the roots are detected. Background seeds are given to voxels having lower intensity than the root positions. The pairwise terms smooth the labeling depending on the gradient values of the image. One difficulty in extracting vessels in the mediastinum is that several neighboring structures are in contact with the pulmonary vessels. In order to prevent over-extracting the ascending aorta and the left ventricles, two landmarks on these structures are detected by the same strategy as the root-position detection and background seeds are set to voxels around the detected landmarks. Another structure that should be separated is the bronchus walls. We also set background seeds around the boundaries of the bronchus regions, which correspond to the bronchus walls, utilizing the automatic bronchus detection described in (Inoue et al. 2013). 

On the other hand, the vessels having tubular appearances in the lung are detected by the multi-scale vessel detector (Kitamura et al. 2012) based on Hessian analysis and machine learning. The method can discriminate vessels accurately in two steps: 
\begin{description}
\item[1)] Estimate the main axis of candidate vessels by the Hessian analysis, 
\item[2)] Discriminate the vessel and the non-vessel based on the Haar-like features extracted from the orientation-normalized local images.
\end{description}

\begin{figure}
\begin{center}
\includegraphics[width=60mm]{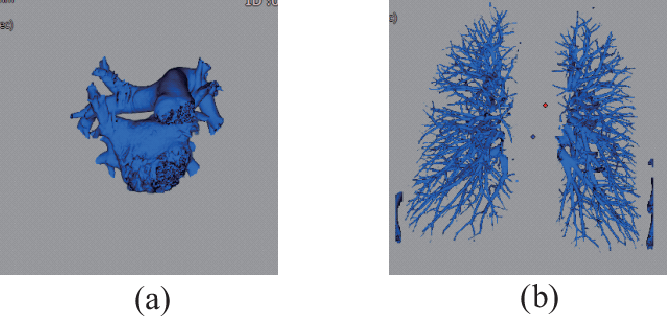}
\end{center}
\caption{Vessel segmentation results (a) in the mediastinum, and (b) in the lungs.}
\label{figVesselSegmentationResults}
\end{figure}

We applied the detector at three scales, using 1.0, 2.0, and 4.0 voxels as the size of the Gaussian kernels for the Hessian. Note that every dataset is rescaled to 0.75mm isotropic voxel size during the vessel extraction and also during the following artery-vein separation, in order to normalize the physical scale. The detected candidates are provided as the foreground seeds to the graph-cut method. The background seeds and the pairwise terms are set in a similar way. Finally, binary segmentation of the entire vessels is obtained as the sum of the results from the mediastinum and the lungs (Fig. \ref{figVesselSegmentationResults}). In order to eliminate false positives, regions that are not connected to the roots are deleted.

\noindent
{\bf Artery-vein separation}\hspace{12pt} Given the root positions and the vessel regions, the third step separates the vessel regions into PA and PV. The energy function consists of unary, pairwise, and higher-order terms:
\begin{equation}
E(X) = \sum_{a \in V} \theta_a(x_a) + \sum_{a \in V, b\in N_a} \theta_{ab}(x_a, x_b) + \sum_{c \in C} \theta_c(X_c),
\label{eqEnergyUPC}
\end{equation}
where $x_a$ takes a binary label \{{\it Artery, Vein}\}. $C$ is a set of cliques, each clique $c$ indexing the DDCP $\theta_c(X_c)$, which is of the form $w_c \cdot \min \left(1, \sum_{a \in c} \frac{1 - x_a}{N} \right)$ (eq. \eqref{eqOrderReduction3}) or $w_c \cdot \min \left(1, \sum_{a \in c} \frac{x_a}{N} \right)$ (eq. \eqref{eqOrderReduction4}) with positive weight $w_c$. Determining $C$ is the most important part of this work, and is discussed in \textsection\ref{subsec_implementation}. The unary terms are set around the artery and vein root positions to force the voxels there to be labeled appropriately according to the following equation:
\begin{equation}
\theta_{a}(x_a) \propto 
\begin{cases}
0 &\text{if } D_{ab} < T_\mathrm{distance},\\
\infty &\text{otherwise},
\end{cases}
\label{eqUnaryForSegmentation}
\end{equation}
where $D_{ab} = \min \left( |P_\mathrm{PA} - P_a|, |P_\mathrm{PV} - P_a| \right)$ is the smaller of the distances from the root positions $P_\mathrm{PA}$ and $P_\mathrm{PV}$ of PA and PV to the voxel $a$ at position $P_a$ and $T_\mathrm{distance}$ is a constant threshold value.

The pairwise terms smooth the labeling between the neighboring voxels (in the 18 neighborhood structure) to different degrees according to the following equation:
\begin{equation}
\begin{split}
& \theta_{ab}(x_a,x_b) \propto \\
& \left \{
%\begin{array}{l}
\begin{split}
&\left( \exp\left(-\frac{G_{ab}^{2}}{\sigma_\mathrm{G}^{2}} \right) + \alpha \right) & &\hspace{-12pt}
  \left( \exp\left(-\frac{H_{ab}^{2}}{\sigma_\mathrm{P}^{2}} \right) + \beta \right)/ D_{ab}^{2}\\
& & &\text{if } x_a \neq x_b,\\
&0, & &\text{otherwise.}
\end{split}
%\end{array}
\right.
\end{split}
\label{eqPairwiseForSegmentation}
\end{equation} 

The function depends on the gradient modulus of the image $G_{ab}$, given by the intensity difference between voxels: $|V_a-V_b|$, and the plateness measure $H_{ab}$ at the mean position of $a$ and $b$ calculated by Hessian analysis (Frangi et al. 1998) to emphasize boundaries where the artery and the vein are in contact. The plateness measure $H_{ab}$ is given by a simplified version of the formulation in (Frangi et al. 1998) :
\begin{equation*}
\label{eqPlateness}
H_{ab} = \exp \left(-\frac{R_\mathrm{AB}}{\sigma_{R_\mathrm{AB}}^{2}} \right) \cdot \left( 1 - \exp \left(-\frac{R_\mathrm{S}^{2}}{\sigma_\mathrm{S}^{2}} \right) \right),
\end{equation*} 
where
\begin{equation*}
\label{eqRinPlateness}
R_\mathrm{AB} = \sqrt{ | \lambda_{1} \lambda_{2} | } / \lambda_{3}, R_\mathrm{S} = \sqrt{\lambda_{1}^{2} + \lambda_{2}^{2} + \lambda_{3}^{2}}.
\end{equation*} 
The $\lambda_1$, $\lambda_2$, and $\lambda_3$ are the three eigenvalues of the Hessian at the voxel. We applied two scales of Gaussian kernels with sizes 0.5 and 1.0 voxels. The plateness measure is calculated from the maximum responses in the two scales. $D_{ab}$ is introduced to prevent shortcuts of label changes near the root positions. The weight parameters were determined empirically, and set as follows:  $T_\mathrm{distance} = 11$mm, $\alpha = 0.01$, $\beta = 0.1$, $\sigma_\mathrm{G} = 50.0$, $\sigma_\mathrm{P} = 0.3$, $\sigma_{R_\mathrm{AB}}= 0.5$, and $\sigma_\mathrm{S} = 8.0$.

\subsection{Implementation details of the data-dependent clique potential}\label{subsec_implementation}
In this section, we describe how the set $C$ of cliques in \eqref{eqEnergyUPC} and the weight $w_c$ for each DDCP $\theta_c(X_c)$ are determined in a principled way in the case of the PA/PV segmentation problem. 

Notwithstanding the strong tendency of the pulmonary vessels to run straight, they of course do not always run in a completely straight line; they sometimes curve or branch. To allow for such flexibility, we encourage sets of voxels forming a curve segment according to its curvature.

To choose the curve segments, we utilize a shortest path algorithm. At each voxel $i$, we construct a shortest-path tree as follows. We first form a graph from the voxels inside the sphere of radius $S/2$ centered at $i$ and the 26-voxel neighborhood topology. Each edge connecting neighboring voxels $a$ and $b$ is given the following weight:
\begin{equation}
E(a, b) = L(a, b)(|V_a - V_b| + \alpha)(|D_a - D_b| + \beta)
\label{eqWeightForShortestPass}
\end{equation} 
where $L(a, b)$ is the physical distance between $a$ and $b$, $V_a$ and $V_b$ are the intensities at $a$ and $b$, $D_a$ and $D_b$ are the distances of $a$ and $b$ from the nearest segmented vessel boundaries, and $\alpha$ and $\beta$ are constant weights. We then find a minimum cost path going through $i$ that connects two voxels on the sphere. The details are as follows:
\begin{description}
\item[1)] Run the Dijkstra algorithm to generate a shortest path tree with $i$ as the root. As a result, every node $j$ has the shortest path $i \rightarrow j$ from the root $i$. Let $c_j$ denote its length, or the sum of the weights of the edges on the shortest path.
\item[2)] For each pair $(j, k)$ of voxels on the sphere, compute the sum of the lengths normalized by the distance between the two voxels: $(c_j + c_k)/|P_j - P_k|$, where $P_j$ denotes the position of the voxel $j$. 
\item[3)] Choose the path $j \rightarrow i \rightarrow k$ with the minimum sum and take it as a curve segment.
\end{description}

This way, we choose one path (curve) for each voxel $i$. The obtained paths tend to be straight avoiding going through a large gap of intensities, and keeping a certain distance from the vessel boundaries. The chosen path constitutes the clique $c$ in \eqref{eqEnergyUPC} and consequently the degree of the higher-order potential $\theta_c(X_c)$ is the number of the voxels therein. The number of voxels is not always the same according to these steps, but it does not become a matter in optimization. Thus, one higher order term per voxel is added by finding the best segment. 

The prediction accuracy of the potentials becomes higher when the ratio of selected cliques lying entirely in one vessel becomes higher. For this reason, the length of the path ($S$) should be larger than the diameter of the vessel so that we can estimate the direction of the vessel. On the other hand, setting a larger number for $S$ increases computational burden. We found the optimal $S$ based on a manually prepared reference segmentation data by running the algorithm on it with varying parameters, and determined $S = 15$ voxel lengths for datasets that were rescaled to 0.75mm isotropic voxel size. We also found the best parameters for eq. \eqref{eqWeightForShortestPass} based on the reference segmentation data by finding the one that resulted in the highest ratio. 

\begin{figure}
\begin{center}
\includegraphics[width=40mm]{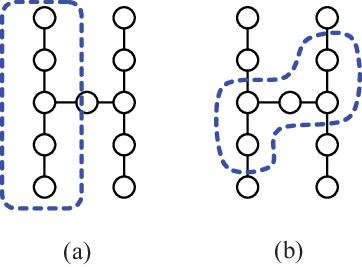}
\end{center}
\caption{Examples of the selected paths (reprinted from Kitamura et al. 2013)}
\label{figExamplesOfPaths}
\end{figure}

Next, for each clique $c$ found this way, we set the weight $w_c$ for the higher-order potential, determining how much we encourage the curve segment to be entirely in one of the segments (artery or vein). Since there is little difference in appearances between artery and vein, here we only distinguish whether all labels are the same or not, regardless of it being artery or vein. This corresponds to giving the same weight for the terms in eqs. \eqref{eqOrderReduction3} and \eqref{eqOrderReduction4}. To determine the weight, the probabilities of the two cases (i.e., the path is entirely in one segment, or not) for paths with varying curvatures are first learned from the reference segmentation data that was manually prepared. For each path, several features are calculated from the voxel set. In the following, we refer to the voxels in a clique (path) in the order from the root to the end: $i \in c = {1,...,n}$ , and each voxel $i$ has the attributes $P_i$ (the 3D position), $V_i$ (the intensity value in HU), and $D_i$ (the distance from the boundaries). The features are:
\begin{description}
\item[i)] the ratio of the length of the path and the straight-line distance between the two endpoints: \\ $\sum_{i \in c} |P_{i+1}-P_i| / |P_n-P_1|$,
\item[ii)] the total curvature along the path: \\ $\sum_{i \in c} |\angle P_{i+1}P_iP_{i-1}|$,
\item[iii)] the maximum curvature on the path: \\ $\max \left(|\angle P_{i+1}P_iP_{i-1}|\right)$,
\item[iv)] the maximum intensity derivative: \\ $\max(V_{i+1}-2 \cdot V_i + V_{i-1})$,
\item[v)] the standard deviation of the intensity derivative: $\sigma(V_i)$,
\item[vi)] the mean derivative of distance from the boundaries: $E(|D_{i+1}-D_{i}|)$, and
\item[vii)] the difference of maximum and minimum distance from the boundaries: $\max(D_i)-\min(D_i)$.
\end{description}
Fig. \ref{figExamplesOfPaths} illustrates two examples of the selected paths. Comparing (a) with (b), the ratio feature i) is larger in (b), as well as the total curvature ii). The maximum intensity derivative takes a large value when a path goes through a boundary of different structures. 

\begin{figure}
\begin{center}
\includegraphics[width=50mm]{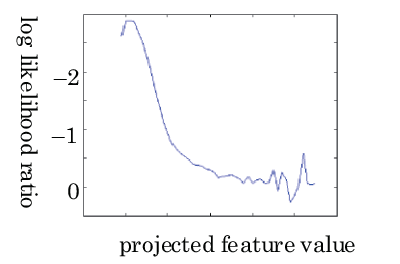}
\end{center}
\caption{Graph of learned log likelihood ratio (reprinted from Kitamura et al. 2013)}
\label{figLikelihoodRatio}
\end{figure}

Then, for each of the two cases (whether all the voxels on the path have the same label or not), the histogram for the feature values is generated and the log likelihood ratio of their probabilities:
\begin{equation*}
-\log(\mathrm{Pr}(\text{\it not all same}) / \mathrm{Pr}(\text{\it all same}))
\end{equation*} 
is learned. To learn the likelihood ratio from the limited number of samples, the feature vectors were projected to one dimension using the Linear Discriminant Analysis before learning. Fig. \ref{figLikelihoodRatio} shows the graph of the learned log likelihood ratio for the projected feature value. The likelihood value corresponding to the feature value for given data is directly used as the weight of the DDCP. Note that the weight $-w_c$ of the energy is clipped to zero when it is positive in order to keep the potential submodular. 

\begin{figure}
\begin{center}
\includegraphics[width=80mm]{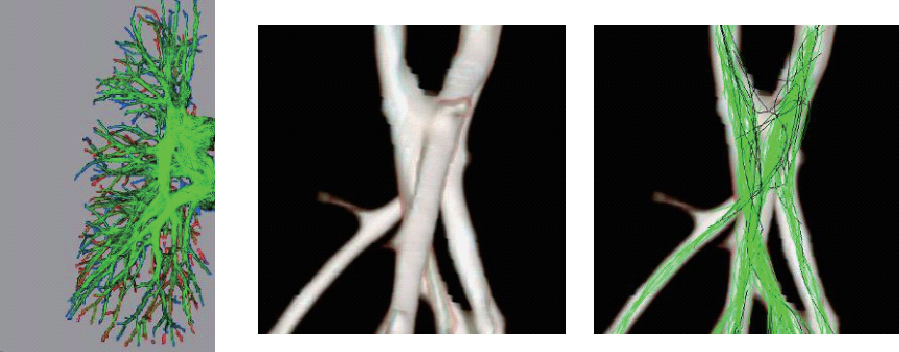}
\end{center}
\caption{Snapshots of the voxel set for data-dependent clique potential. Each green filament represents a higher order clique. The potential weight $w_c$ for the clique c is determined according to its shape, in this case curvature: the higher the weight is, the brighter it is drawn. (reprinted from Kitamura et al. 2013)}
\label{figSnapshotsOfASPP}
\end{figure}

Snapshots of selected segments for the DDCP are shown in Fig. \ref{figSnapshotsOfASPP}. Each segment is drawn in green through the selected voxels. The brightness indicates the likelihood: the higher the likelihood of the segment is, the brighter it is drawn. 

\subsection{Incorporating spatial arrangement features into data-dependent clique potential}\label{subsec_spatial}
Although there are little difference between the appearances of artery and vein, it is known that an artery often go along an airway. Here we consider giving different weights for the DDCP depending on the proximity to the airways. This improvement relies on the approach in (Mekada et al. 2006), which classifies vessels based on two anatomical features; the distance from the bronchus region to the vessel segment ($D_b$) and the distance between the nearest interlobar to the vessel ($D_v$). The interlobar is approximated by a 3D extended Voronoi diagram of the bronchus tree. Since the PA runs parallel to the bronchi, the $D_b$ for the PA branches will be smaller than those of the PV. On the other hand, the $D_v$ for the PV branches tend to be smaller than those of PA, because they run near the interlobar of the lung segment. From these observations, segmented vessels can be classified by an A/V classification measure:
\begin{equation}
\arctan(D_b' / D_v')
\label{eqDbDvRatio}
\end{equation}
where
\begin{equation*}
D_b'=\frac{1}{|c|}\sum_{i \in c}D_b(i)/\sigma_{D_b},
D_v'=\frac{1}{|c|}\sum_{i \in c}D_v(i)/\sigma_{D_v}.
\end{equation*}
The parameters $\sigma_{D_b}$ and $\sigma_{D_v}$ represent the standard deviations of the values $D_b$ and $D_v$ in the input volume; $c$ is the group of voxels to be evaluated, which are the voxels in the selected clique in our case. 

\begin{figure}
\begin{center}
\includegraphics[width=80mm]{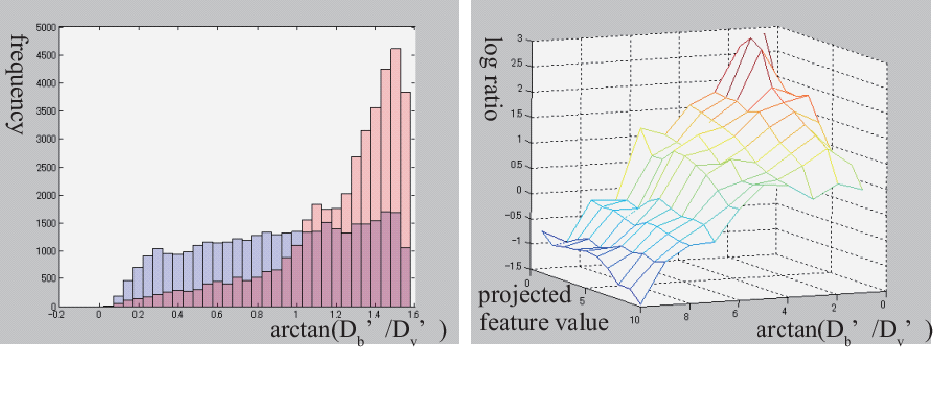}
\end{center}
\caption{(LEFT) One-dimensional histograms that voxels in a clique were all artery or all vein, where blue is vein and pink is artery, (RIGHT) the log ratio for the artery-ness and vein-ness for the projected feature value (described in 3.2) and the artery-vein classification measure}
\label{figHistogram}
\end{figure}

We derive a two-dimensional histogram for the projected feature value described in \textsection\ref{subsec_implementation} (which is a measure that a clique has a straight shape) and the A/V classification measure from the training datasets. And the log likelihood ratio of the artery-ness (all voxels in a clique are artery) given by 
\begin{equation*}
-\log(\mathrm{Pr}(\text{\it not all same}) / \mathrm{Pr}(\text{\it all artery}))
\end{equation*} 
and the vein-ness (all voxels in a clique are vein) given by 
\begin{equation*}
-\log(\mathrm{Pr}(\text{\it not all same}) / \mathrm{Pr}(\text{\it all vein}))
\end{equation*} 
are learned. Fig. \ref{figHistogram}(a) is one dimensional histograms that voxels in a clique were all artery or all vein. Fig. \ref{figHistogram}(b) is the log ratio for the artery-ness and vein-ness. We can see that a higher-order clique tend to be entirely artery when it has a straight shape and close to the airway. Meanwhile a clique can be vein, regardless of the two features. Each of the two likelihoods gives the weights for the terms in eqs. \eqref{eqOrderReduction3} and \eqref{eqOrderReduction4}. Note that, in the following verification tests, we employed the fully-automatic results obtained by the airway extraction method (Inoue et al. 2013) which has been demonstrated to have high extraction performance among the state of the arts methods by the public benchmarking {\it EXACT09} (Lo et al. 2012).

\section{Experimental Validation}\label{sec_experiments}
Here, we describe both quantitative and qualitative evaluations to show the efficacy of the proposed method.

\subsection{Quantitative evaluation}\label{subsec_quantitative}
For quantitative validation, we compare our method with the method in (Mekada et al. 2006), which is the only fully-automatic method for PA/PV segmentation we could find. We used ten chest CT Angiography images that were not used for learning. The datasets including thoracic regions were imaged using Toshiba Medical Systems multi-detector CT scanner at $120$kVp. The product of exposure time and tube current ($mAs$) was 159 on average (Min: 103, Max: 218). The images were acquired at $1.0$mm slice thickness and reconstructed with $0.8$mm slice thickness and $0.67 \times 0.67$mm$^2$ in plane resolution. All the patients in the datasets are known to suffer some lung disease and at least one tumor exists in the lungs per dataset. 

Ground-truth data were prepared for these images by manually labeling the artery and vein regions. Note that the ground-truth data was established only for vessels with CT values more than $-200$HU in the lungs. Even under this condition, we consider this ground-truth data to be sufficient to validate A/V separation, since it covers most important vessels except for peripheral branches which are not often in contact with other vessels. The ground-truth data is directly used for calculating volume-based measures. The volume based classification accuracies were calculated as the percentages of the volume that the vessel in the ground truth was correctly classified in the segmentation results.

Also, since the volume-based measures are biased by the proximal vessels' much larger volume compared to the peripheral vessels', we also evaluated the performance by length-based measures. The artery and vein centerlines were generated by thinning operation, which can prevent generating spike branches or holes, to each of the labeled regions of PA and PV. We confirmed that the resulting centerlines were going through the vessel centerlines by visual inspection. The length based classification accuracies were calculated as the percentages of the length that the vessel in the ground truth was correctly classified in the segmentation results, in the same way as volume-based accuracies. 

For each of the volume- and length-based measures, three kinds of the accuracy measures were calculated, corresponding to the case when the ground truth included only PA, only PV, and both (PA$\cup$PV). Miss-extraction rates were also calculated as the percentages of the vessel that was not segmented (extracted) as artery or vein. 

\begin{figure}
\begin{center}
\includegraphics[width=80mm]{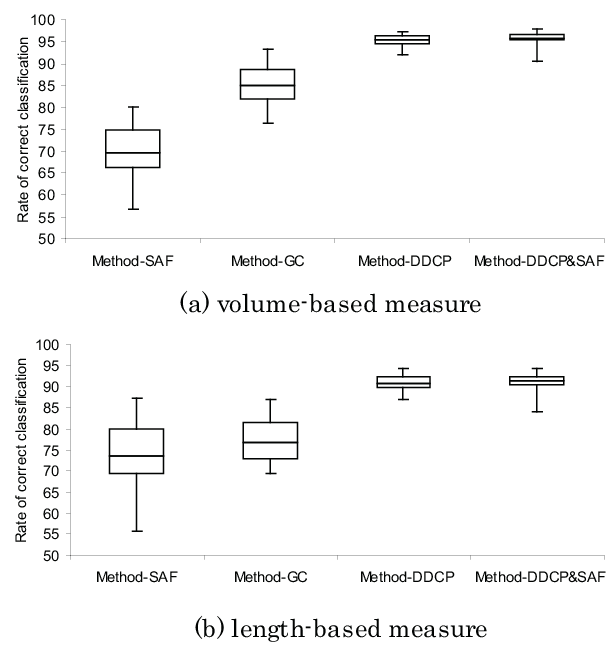}
\end{center}
\caption{Box plots of (a) volume-based and (b) length-based correct classification rates of PA$\cup$PV comparing four methods. From left to right of the horizontal axis, Method-SAF: the classification method using spatial arrangement features. Method-GC: graph cuts without the DDCP, Method-DDCP: graph cuts with the DDCP, and Method-DDCP\&SAF: graph cuts combining the DDCP and the spatial arrangement feature.}
\label{figBoxPlots}
\end{figure}

We compared four types of extraction method: Method-SAF) the classification method in (Mekada et al. 2006) using spatial arrangement features, Method-GC) first-order graph cuts without the data-dependent clique potential, Method-DDCP) higher-order graph cuts with the DDCP, and Method-DDCP\&SAF) higher-order graph cuts with the DDCP with the spatial arrangement feature. Method-SAF is an exact replication of (Mekada et al. 2006), which classifies the type of vessels in the following steps.
\begin{description}
\item[1)] The tree structure of the branches is obtained by thinning operation for the segmented vessel regions.
\item[2)] The branches in the tree are merged to groups considering the connection relationship and contacting points of PA and PV.
\item[3)] Each group is classified into artery or vein by thresholding the A/V classification measure:
\begin{equation}
\mathit{Type} = 
\begin{cases}
\mathit{artery}, &\text{if } \arctan(D_b'/D_v')<0.5, \\
\mathit{vein}, &\text{otherwise}
\end{cases}
\label{eqAVClassification}
\end{equation} 
\item[4)] Segmented volumes of PA and PV are generated by dilating the classified branches until they conflict with the other label. \end{description}
Method-GC is for giving a baseline of conventional first-order graph cuts approach. Method-DDCP and DDCP\&SAF are the approaches proposed in this paper, corresponding to the ones described in \textsection\ref{subsec_implementation} and \textsection\ref{subsec_spatial}, respectively. With regard to the robust $P^n$ Potts model, we set $N = 3$ for both of \eqref{eqOrderReduction3} and \eqref{eqOrderReduction4} during these evaluation, since it generated slightly better segmentation than $N = 1$. All steps were executed automatically without any user interaction. Tables \ref{tabVolume} and \ref{tabLength} show the volume- and length-based measurement results, respectively. To summarize the results, the average rate of length-based correct classification of PA$\cup$PV were 73.6\% in the case of Method-SAF, 77.6\% in the case of Method-GC, 90.8\% in the case of Method-DDCP, and 91.0\% in the case of Method-DDCP\&SAF. The average miss-extraction rate was 3.3\%, which was common between all methods as they relied on the same vessel-extraction method. The box plot of the results is shown in Fig. \ref{figBoxPlots}.

\begin{figure*}
\begin{center}
\includegraphics[width=150mm]{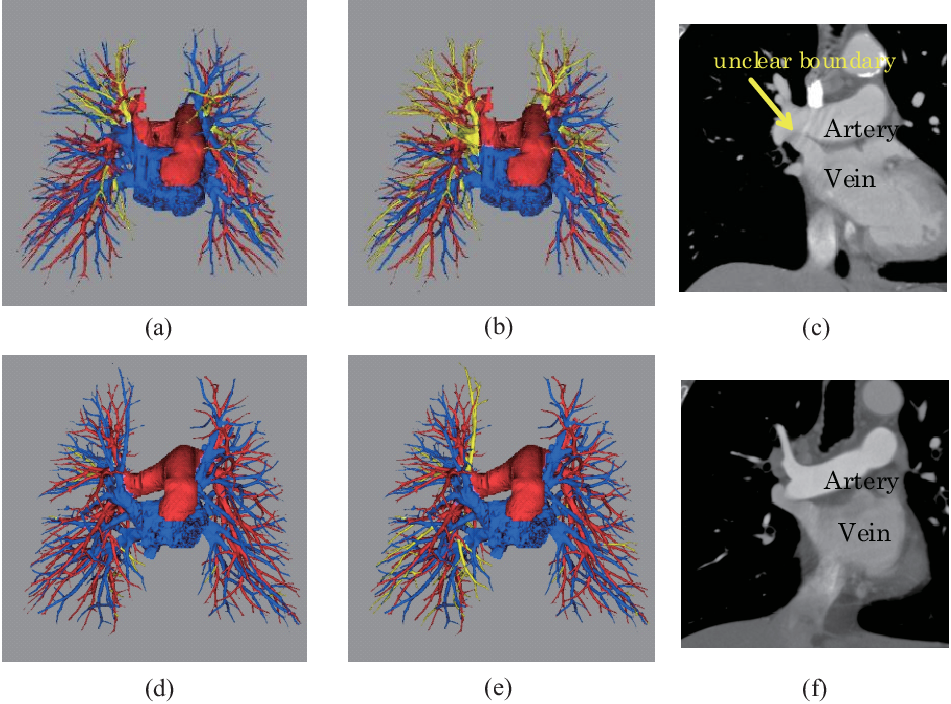}
\end{center}
\caption{The segmentation results of the case that the artery and the vein are in contact over a large area. (a)-(c): the case with an unclear boundary. (d)-(f): the case where the difference of intensity values between the artery and the vein are comparably higher. (a) and (d) were obtained from the method with DDCP. (b) and (e) are the baseline cases without DDCP. The red and blue regions represent artery and vein, respectively. The yellow represents misclassified regions. (c) and (f) are coronal images around the pulmonary hilum. (reprinted from Kitamura et al. 2013)}
\label{figSegmentationResults}
\end{figure*}

\begin{table*}
% table caption is above the table
\caption{Volume-based correct classification rates (\%) of PA and PV for each case}
%\caption{Please write your table caption here}
\label{tabVolume}
\begin{tabular}{|c|c|c|c|c|c|c|c|c|c|c|}
\hline
\multicolumn{1}{|c|}{Case} & \multicolumn{2}{|c|}{Method-SAF} & \multicolumn{2}{|c|}{Method-GC} & \multicolumn{2}{|c|}{Method-DDCP} & \multicolumn{2}{|c|}{Method-DDCP\&SAF} & \multicolumn{2}{|c|}{Miss} \\
\hline
No & PA & PV & PA & PV & PA & PV & PA & PV & PA & PV \\
\hline
\hline
1 & 80.7 & 80.8 & 91.2 & 91.0 & 97.2 & 94.2 & 97.4 & 93.4 & 0.8 & 0.3 \\
\hline
2 & 76.8 & 69.1 & 94.9 & 68.2 & 93.6 & 96.8 & 95.1 & 96.2 & 0.6 & 0.7 \\
\hline
3 & 69.8 & 69.3 & 92.4 & 73.1 & 92.4 & 93.5 & 94.9 & 93.7 & 2.0 & 1.7 \\
\hline
4 & 71.9 & 81.2 & 93.4 & 90.4 & 97.8 & 95.6 & 99.2 & 96.0 & 0.4 & 1.0 \\
\hline
5 & 62.5 & 71.7 & 96.0 & 73.8 & 97.4 & 96.6 & 98.0 & 98.0 & 0.9 & 1.4 \\
\hline
6 & 64.7 & 75.1 & 90.3 & 72.2 & 92.7 & 95.2 & 92.8 & 97.9 & 0.7 & 0.9 \\
\hline
7 & 80.1 & 74.3 & 92.4 & 94.0 & 96.5 & 97.7 & 95.7 & 97.8 & 0.8 & 1.6 \\
\hline
8 & 60.1 & 72.2 & 90.6 & 70.6 & 96.6 & 94.1 & 97.9 & 94.3 & 1.2 & 1.1 \\
\hline
9 & 35.7 & 83.8 & 89.3 & 59.9 & 89.3 & 95.7 & 89.3 & 92.3 & 2.2 & 1.3 \\
\hline
10 & 56.7 & 66.4 & 95.9 & 75.2 & 97.2 & 93.8 & 97.4 & 95.9 & 0.6 & 1.5 \\
\hline
\hline
Mean & 65.9 & 74.4 & 92.6 & 76.9 & 95.1 & 95.3 & 95.8 & 95.6 & 1.0 & 1.2 \\
\hline
\end{tabular}
\end{table*}

\begin{table*}
% table caption is above the table
\caption{Length-based correct classification rates (\%) of PA and PV for each case}
%\caption{Please write your table caption here}
\label{tabLength}
\begin{tabular}{|c|c|c|c|c|c|c|c|c|c|c|}
\hline
\multicolumn{1}{|c|}{Case} & \multicolumn{2}{|c|}{Method-SAF} & \multicolumn{2}{|c|}{Method-GC} & \multicolumn{2}{|c|}{Method-DDCP} & \multicolumn{2}{|c|}{Method-DDCP\&SAF} & \multicolumn{2}{|c|}{Miss} \\
\hline
No & PA & PV & PA & PV & PA & PV & PA & PV & PA & PV \\
1 & 87.2 & 87.5 & 85.1 & 84.5 & 94.5 & 90.5 & 94.8 & 89.5 & 2.3 & 1.8 \\
\hline
\hline
2 & 84.0 & 74.3 & 88.4 & 62.2 & 87.7 & 92.9 & 89.0 & 91.5 & 2.9 & 2.5 \\
\hline
3 & 70.2 & 72.0 & 84.5 & 60.8 & 86.0 & 87.9 & 89.4 & 87.7 & 5.2 & 4.7 \\
\hline
4 & 80.8 & 85.1 & 87.0 & 81.8 & 94.9 & 90.1 & 97.4 & 90.9 & 2.0 & 4.7 \\
\hline
5 & 66.7 & 68.9 & 89.8 & 68.2 & 93.2 & 92.7 & 94.0 & 94.5 & 3.4 & 3.7 \\
\hline
6 & 68.6 & 81.6 & 84.7 & 61.3 & 88.3 & 91.6 & 88.5 & 94.8 & 2.2 & 2.6 \\
\hline
7 & 87.8 & 72.5 & 86.3 & 87.9 & 93.9 & 94.5 & 91.9 & 94.6 & 2.4 & 4.1 \\
\hline
8 & 66.1 & 78.1 & 85.6 & 60.5 & 91.9 & 87.4 & 94.4 & 87.7 & 3.2 & 4.2 \\
\hline
9 & 34.0 & 80.3 & 85.4 & 51.1 & 86.1 & 92.0 & 80.6 & 89.0 & 3.9 & 3.5 \\
\hline
10 & 57.5 & 70.3 & 92.1 & 62.6 & 94.8 & 87.6 & 95.2 & 90.8 & 1.5 & 4.6 \\
\hline
\hline
Mean & 70.3 & 77.1 & 86.9 & 68.1 & 91.1 & 90.7 & 91.5 & 91.1 & 2.9 & 3.6 \\
\hline
\end{tabular}
\end{table*}

\begin{figure}
\begin{center}
\includegraphics[width=70mm]{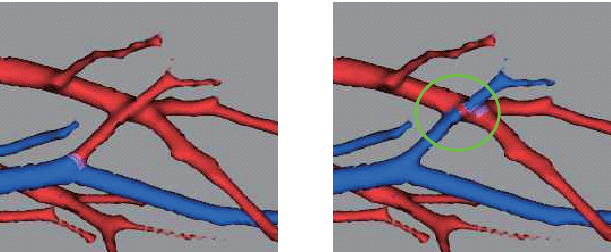}
\end{center}
\caption{The comparison of segmentation results when (LEFT) $N = 1$ and (RIGHT) $N = 3$. On the right image, the voxels along the vein branch was allowed to have different label.}
\label{figEffectOfRobustModel}
\end{figure}

Firstly, let us discuss the impact of the DDCP by comparing Method-GC and Method-DDCP. Shown in Fig. \ref{figSegmentationResults}(a), (b), (c) is an example of the case where a major difference was seen between the two methods. Method-GC, the baseline, generated a large misclassified region, as it used only the pairwise term as a typical conventional method does. Such a method tends to fail to separate regions that are in contact over a large area or with an unclear boundary (Fig. \ref{figSegmentationResults}(c)). The case where even Method-GC achieved a high rate of correct classification is shown in Fig. \ref{figSegmentationResults}(d), (e), (f). In this case, the difference of intensity values between the artery and the vein were comparably higher due to the timing of the injection of the contrast agents (Fig. \ref{figSegmentationResults}(f)). However, Method-DDCP obtained a higher rate, correctly classifying more vessels around the peripheral.Fig. \ref{figEffectOfRobustModel} shows an efficacy of the robust $P^n$ Potts model. The peripheral branch indicated by the circle was successfully segmented when $N = 3$, while in contrast the method using $N = 1$ failed.

\begin{figure}
\begin{center}
\includegraphics[width=70mm]{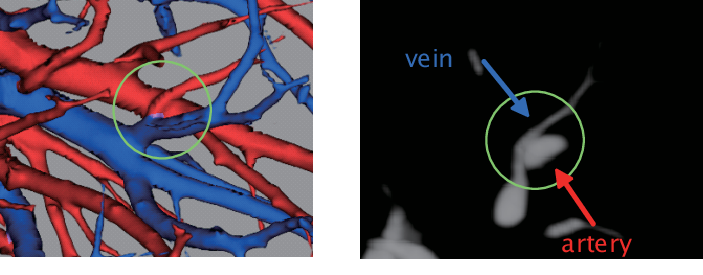}
\end{center}
\caption{A failure case of the method with the DDCP. (LEFT) 3D rendering of the miss-recognized point, (RIGHT) a cross section image of the contacting point, The vein branch was classified as artery.}
\label{figFailureCase}
\end{figure}

A failure case of Method-DDCP is shown in Fig. \ref{figFailureCase}. Typically, the proposed method fails where curving or branching vessels are in contact. Because the method assumes that pulmonary vessels run straightly, the assumption does not meet this situation. The average processing time of the Method-GC and Method-DDCP were 52.5 seconds and 93.2 seconds per dataset on a quad-core 2.8GHz PC.

\begin{figure}
\begin{center}
\includegraphics[width=70mm]{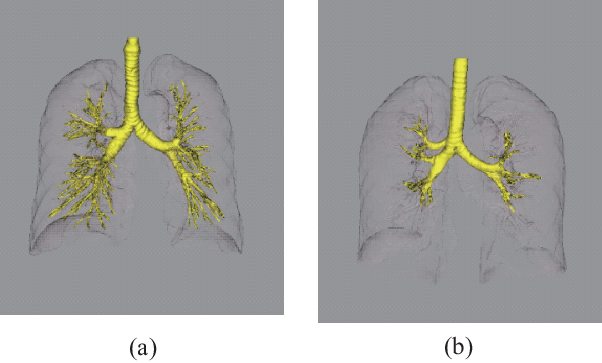}
\end{center}
\caption{The automatic bronchus extraction results used for artery-vein separation in the qualitative evaluation. (a) is the extracted bronchus when the classification accuracy was high, (b) is the one when the classification accuracy was low.}
\label{figBronchusResults}
\end{figure}

Next, let us show the efficacy of the spatial arrangement features. The correct classification rate of Method-SAF was 73.6\%, although it is reported in (Mekada et al. 2006) that the method achieved 87\% in the three cases with complete bronchus trees. Similar rates were obtained in the instances of the testing datasets that ranked high, but the rates for the rest of the datasets were much lower. The reason is that we cannot always obtain automatic bronchus extraction results good enough for artery-vein classification. As the two examples in Fig. \ref{figBronchusResults} show, the classification performance significantly decreases in the cases where the extracted bronchus tree does not extend to the peripheral in the lungs. This is mainly caused not by the limitation of the extraction performance but by the fact that bronchus cannot always be seen clearly, depending on the patient. The Method-DDCP\&SAF achieved the highest classification rate of 91.0\% on average. In a few cases, however, the rates are much worse compared to Method-DDCP. The reason for the diminished performance was also the limited extraction performance of the bronchus. Improving its robustness is an important issue to be solved in the future.

\subsection{Subjective tests in clinical settings}\label{subsec_subjective}
Now we describe the details of the subjective tests in clinical settings. Since these tests were done at a hospital where the proposed method has been deployed as commercial software, there were no conflicts of interest between the clinicians that conducted the tests and the authors of this paper. The thoracic regions were imaged using Toshiba Medical Systems multi-detector CT scanner at $120$kVp under automatic exposure control. The images were acquired at $0.5$mm slice thickness and reconstructed with $0.5$mm slice thickness and $0.5 \times 0.5\text{mm}^2$ in plane resolution. As described below, test datasets were acquired at several different image contrasts. Contrast materials were injected into patients at $1.2-1.5$ml/sec. for obtaining low contrast data or $3.0$ml/sec. for obtaining high contrast data.

\begin{figure}
\begin{center}
\includegraphics[width=50mm]{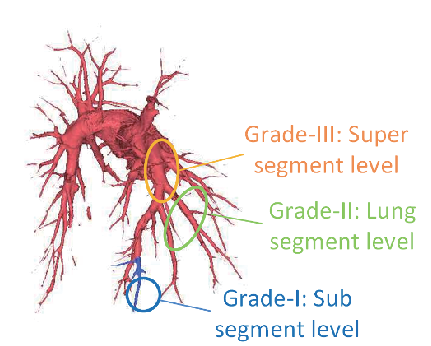}
\end{center}
\caption{Example of the three grades of misclassified branches in the subjective tests.}
\label{figExampleOfGrades}
\end{figure}

The procedure and viewpoints of these tests are as follows. Misclassified branches were detected by eye-balling by three experts and classified into three grades by consensus. The three grades were determined by the depth of the most proximal misclassified branch. Because the pulmonary vessels have tree structures, total misclassified region grows larger as the misclassified position becomes proximal. As is shown in Fig. \ref{figExampleOfGrades}, Grade-II means that the misclassified branches form an equivalent of one lung-segment; it is known that both of the human lungs are classified into about ten such segments. In a relative manner, Grade-I and Grade-III represent the cases where misclassified branches range less than one lung segment and more than one lung-segment, respectively.

Testing datasets were categorized into three groups: the low contrast, the high contrast, and the mixture group. The images in the low-contrast group have less than $200$HU around both of the PA and PV trunk regions. The images in the high-contrast group have more than $200$HU. The images in the mixture group have more than $200$HU around one of the PA or PV trunks. The subjective tests were done at slightly different settings: the robust $P^n$ Potts model was not used for Method-DDCP (i.e. $N = 1$).

\subsubsection{Subjective test - I}
One of the two tests is for evaluating the efficacy of the higher-order potentials. Similar to the previously-mentioned quantitative evaluation, two extraction results obtained by the energy functions with and without the DDCP (Method-GC and DDCP) were compared. The numbers of cases in the high-contrast, low-contrast, and mixture groups were 10, 9, and 5, with the total of 24 cases used. 

The summary of this test is shown in Table 3. The total number of misclassifications using Method-DDCP decreased by 52\% compared to Method-GC. It is especially notable that the number of Grade-III misclassification dropped by two thirds, showing that the DDCP greatly contributes to successful recognition of large amount of branches. These results agree well with the quantitative evaluation and prove the robustness of the method with a large number of datasets.

\begin{table}[h]
\caption{Total number of misclassified branches in 24 cases obtained by Method GC and DDCP}
\label{tabSubjective1}
\begin{tabular}{|c|c|c|c|c|}
\hline
 & Grade-III & Grade-II & Grade-I & Total \\
\hline
\hline
Method-GC & 33 & 58 & 63 & 154 \\
\hline
Method-DDCP & 11 & 41 & 22 & 74 \\
\hline
\end{tabular}
\end{table}

\subsubsection{Subjective test - II}
We also assessed the dependence of the methods on the degree of contrast enhancement. The numbers of cases in the high-contrast, low-contrast, and mixture groups were 16, 13 and 11, with the total of 40 cases used. Only the method with the DDCP (Method-DDCP) was evaluated in this test. Note that adding contrast at more than $200$HU in angiographic imaging is considered a standard protocol; thus this evaluation is for testing robustness against contrast variability in a clinical setting.

\begin{figure}
\begin{center}
\includegraphics[width=80mm]{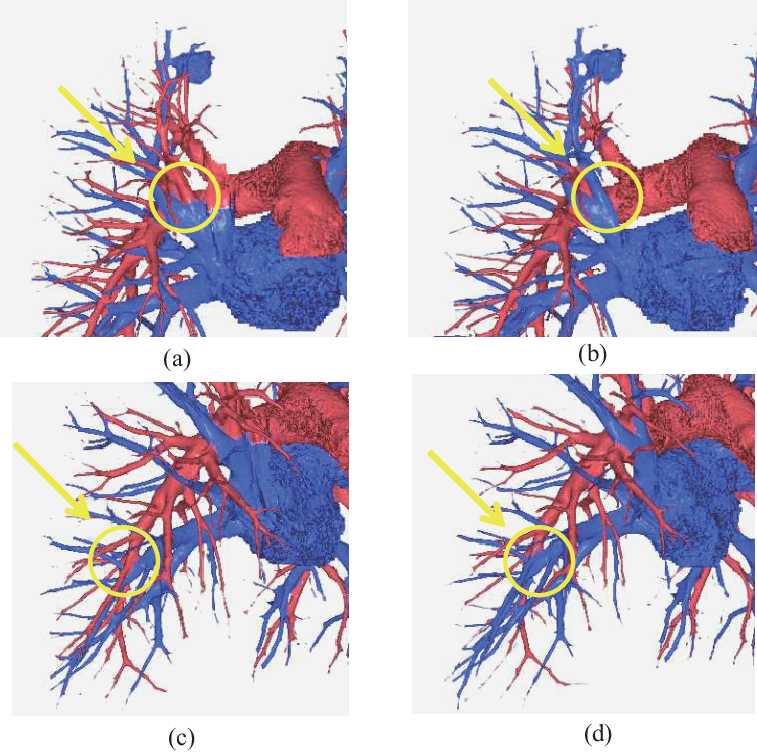}
\end{center}
\caption{The segmentation results for two cases acquired from the same patient but adding different degrees of contrast. (a) and (c) are the results in the high contrast group. (b) and (d) are in the mixture group (lower contrast enhancement). The branches around superior vena cava (SVC) were classified better in (b) than in (a), while the branches in the lower lobe were classified better in (c) than in (d).}
\label{figResultsForDifferentContrast}
\end{figure}

The summary of evaluation results is shown in Table \ref{tabSubjective2}. The method achieved higher classification accuracies as the contrast was enhanced more. The method generated an extremely low rate of misclassification (0.19 at Grade-III, 1.19 at Grade-II on average) in the high contrast group. One exception is that some segmentation results in the high contrast group were worse than in the low contrast group, especially around the superior vena cava (SVC), due to the artifacts caused by the contrast agent. The examples that were acquired from the same patient but adding different degrees of contrast are shown in Fig. \ref{figResultsForDifferentContrast}. To summarize, the method generated almost no major failure when an appropriate contrast enhancement is given, and also do not require severe control of the timing of contrast agent injection. 

\begin{table}[h]
\caption{Total number of misclassified branches in 40 cases obtained by Method DDCP for datasets having different contrast}
\label{tabSubjective2}
\begin{tabular}{|c|c|c|c|c|}
\hline
 & Grade-III & Grade-II & Grade-I & Total \\
\hline
\hline
High contrast & 3 & 19 & 16 & 38 \\
\hline
Mixture & 5 & 14 & 12 & 31 \\
\hline
Low contrast & 10 & 29 & 16 & 55 \\
\hline
\end{tabular}
\end{table}

\section{Discussion}\label{sec_discussion}
It should be noted that the separation performance is affected by the image resolution. The vessels with diameters that are less than the resolution scale are not separated well in principle as is described in 3. In addition, as the ground-truth data was prepared only for relatively large vessels, such small vessels were not evaluated in our study. Processing with the increased resolution would be a straightforward way to raise the performance, but it also increases computational complexity and consumed memory size. We designed the experimental settings balancing these factors to meet our purpose of pre-surgery simulator. Dealing with small vessels and more comprehensive evaluation would be a future work.

Although a few promising methods have been proposed for PA/PV segmentation, we could not directly compare our method with them, due to the differences in the prerequisite information/user intervention in each method. Since PA/PV segmentation consists of several tasks such as seed selection, vessel segmentation, and artery-vein separation, matching the condition of each step cannot be done easily. Still, to put our result in perspective, let us compare our qualitative results with state-of-the-art methods in related but not identical settings. The method in (Saha et al. 2010) achieved 95\% correct classification rate on two (non-angiography) CT datasets, using 25-40 seeds specified by the user, whereas it is reported that (Mekada et al. 2006), using anatomical knowledge on pulmonary vessels and airway, achieved 87\% on three CT datasets, given airway-lung segmentation a priori. Since these methods require different input (image modality, seeds, a priori segmentation), it is not possible to directly compare them with our method. Nevertheless, we consider our method highly competitive as it achieves 90.7\% without any user interaction. Moreover, the computational time is short enough for practical use.

Next, we discuss the features we use in this study to determine the weights of the DDCPs. In the case of lower-order clique potentials, there are well-studied features like the correlation coefficient (Lei 2010), which is parameterized by the image intensity and the distance between voxels. However, in this paper we focus on the relationships between much more numerous voxels and consider the shapes they form. Although we learned the relationship between the features and the weights, we determined some parameters heuristically in choosing the features itself. Thus, the exploited features here might not be optimal. A generalized scheme to design feature vectors belongs to a future work. In this paper we verified that, once the features are chosen, giving the weights of the DDCPs based on the statistics of the reference data is effective in obtaining promising results.

Regarding the subjective tests, the experts who performed the tests concluded that the proposed method was effective for reducing human workload in generating 3D images of PA and PV. An alternative mean to obtain the 3D images at clinical practice is taking two-phase images adding different contrast for PA and PV. Using the images acquired in such a manner, PA and PV can be visualized by the conventional volume rendering method. However, they commented that the segmentation accuracy of the fully automated method presented here was clinically acceptable and could replace the current two-phase imaging technique. Furthermore, the proposed method also has an impact in reducing the patient dose, as the method requires only one phase.

\begin{figure}
\begin{center}
\includegraphics[width=80mm]{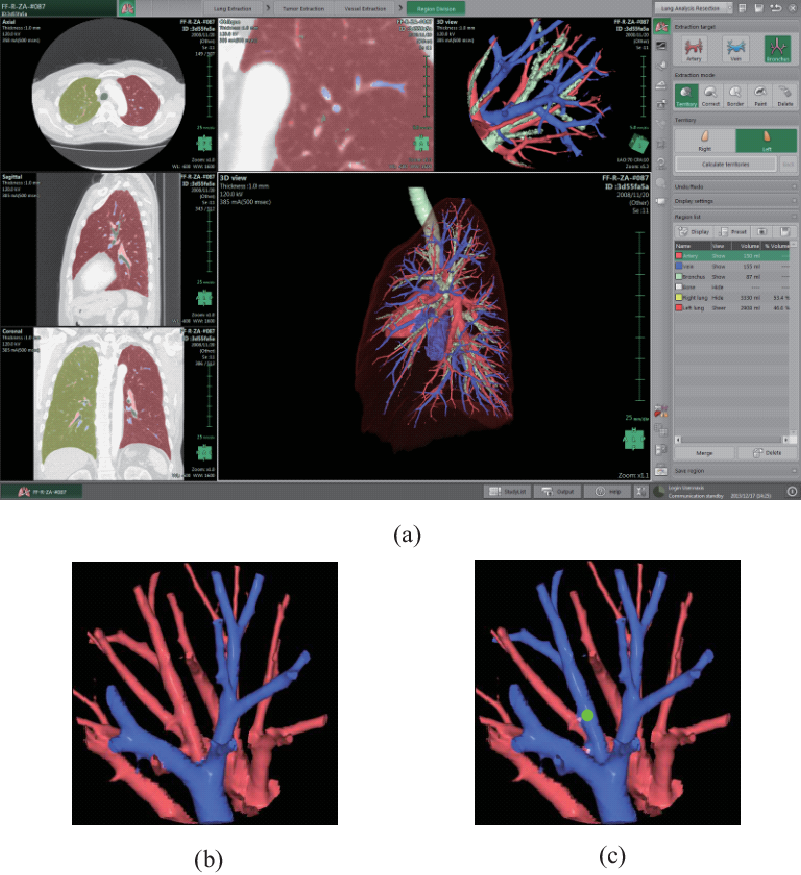}
\end{center}
\caption{The pre-surgery simulation system in which the pulmonary artery-vein segmentation method and its interactive modification function have been implemented. (a) The entire view of the graphical user interface. The 3D rendering image is shown at the center. A user can input 3D points to modify the segmentation results through the interface. (b) The pulmonary artery-vein segmentation result before modification. (c) The segmentation result after modification. The green dot represents the inputted seed point.}
\label{figPresurgerySimulator}
\end{figure}

Another remarkable feature of the proposed method is its quick response to user operation. When additional inputs are given to modify unsatisfactory segmentation results of PA and PV, optimal solution can be computed quickly by using the flow-reusing techniques (Kohli et al. 2005; Boykov and Kolmogorov 2004), while keeping the strong tendency that the segmentation result keeps straight configuration. We developed the pre-surgery simulation system in which the proposed method and a graphical user interface system for modifying results were implemented (Fig. \ref{figPresurgerySimulator}). By using the flow-reusing technique, the system realizes a response within one second upon user interaction. Surgeons who use our pre-surgery simulation system commented that the procedures for further modification by a thoracic surgeon, taking anatomical variability into account, took only 5-10 minutes per patient (Saji et al. 2013). The system is now being used routinely in the institution as is mentioned in the paper.

Although the number of existing methods that utilize higher-order potentials is still limited, we proposed a novel approach to effectively exploit the prior knowledge of the object shape. The key idea is selecting voxels in a clique to represent at least a part of the object shape. As an embodiment, we present an effective modeling of vessels. The verification tests were done especially for PA and PV, but this approach can be immediately applicable for the vessels in other parts of the body such as head and abdomen. As the basic idea is general, considering other voxel selection approach as per the shapes such as plate-like or arc-like structures is likely to improve the segmentation accuracy of other kinds of objects.

\section{Conclusion}\label{sec_conclusion}
We proposed a novel segmentation method that utilizes higher-order functions which allow modeling the shape of segments such as the complex anatomy of pulmonary vessels. The higher-order terms encourage sets of pixels to be entirely in one segment or the other, and they can be converted into submodular first-order terms so that it can be globally minimized. The key feature of the proposed method is selecting pixels in a clique according to the shape to be segmented. We presented a fully-automatic pulmonary artery-vein segmentation method. The verification tests showed that the method achieved clinically acceptable accuracies. We consider this method applicable to various other segmentation problems.

% For one-column wide figures use
%\begin{figure}
% Use the relevant command to insert your figure file.
% For example, with the graphicx package use
%  \includegraphics{example.eps}
% figure caption is below the figure
%\caption{Please write your figure caption here}
%\label{fig:1}       % Give a unique label
%\end{figure}
%
% For two-column wide figures use
%\begin{figure*}
% Use the relevant command to insert your figure file.
% For example, with the graphicx package use
%  \includegraphics[width=0.75\textwidth]{example.eps}
% figure caption is below the figure
%\caption{Please write your figure caption here}
%\label{fig:2}       % Give a unique label
%\end{figure*}
%
% For tables use
%\begin{table}
% table caption is above the table
%\caption{Please write your table caption here}
%\label{tab:1}       % Give a unique label
% For LaTeX tables use
%\begin{tabular}{lll}
%\hline\noalign{\smallskip}
%first & second & third  \\
%\noalign{\smallskip}\hline\noalign{\smallskip}
%number & number & number \\
%number & number & number \\
%\noalign{\smallskip}\hline
%\end{tabular}
%\end{table}

\begin{acknowledgements}
%If you'd like to thank anyone, place your comments here
%and remove the percent signs.
We appreciate the advice regarding clinical knowledge and subjective tests given by the Department of Radiology, NTT East Sapporo Hospital. This research was done in cooperation with the Department of Thoracic Surgery, Tokyo Medical University Hospital. In the course of this work, Hiroshi Ishikawa was partially supported by the Grant-in-Aid for Scientific Research on Innovative Areas \#26108003 from the Japan Society for the Promotion of Science (JSPS) as well as the CREST grant from the Japan Science and Technology Agency (JST).
\end{acknowledgements}

% BibTeX users please use one of
%\bibliographystyle{spbasic}      % basic style, author-year citations
%\bibliographystyle{spmpsci}      % mathematics and physical sciences
%\bibliographystyle{spphys}       % APS-like style for physics
%\bibliography{}   % name your BibTeX data base

% Non-BibTeX users please use

\end{document}